\documentclass[letterpaper, 10 pt, conference]{ieeeconf}  

\IEEEoverridecommandlockouts                              

\usepackage{graphicx} 
\usepackage{amsmath}
\usepackage{amssymb}
\usepackage{bbold}
\usepackage{eqnarray}
\usepackage{bbm}
\usepackage{xcolor}
\usepackage{url}
\usepackage{hyperref}

\newcommand{\ONE}{\mathbb{1}}

\newcommand{\RR}{\mathbb{R}}
\newcommand{\BB}{\mathbb{B}}
\newcommand{\Tau}{\mathbb{T}}

\newcommand{\strictorder}{\prec}

\newtheorem{definition}{\textbf{Definition}}

\newtheorem{lemma}{\textbf{Lemma}}
\newtheorem{corollary}{\textbf{Corollary}}
\newtheorem{example}{\textbf{Example}}

\title{\LARGE \bf
Hierarchical Potential-based Reward Shaping from Task Specifications
}

\author{
Luigi Berducci\textsuperscript{*\rm 1},
Edgar A. Aguilar\textsuperscript{*\rm 2},
Dejan Ni\v{c}kovi\'c\textsuperscript{\rm 2},
Radu Grosu\textsuperscript{\rm 1}
\thanks{*Indicates authors with equal contributions}
\thanks{$^{1}$TU Wien, Cyber-Physical Systems Group}%
\thanks{$^{2}$AIT Austrian Institute of Technology GmbH}%
}

\begin{document}

\maketitle
\thispagestyle{empty}
\pagestyle{empty}

\begin{abstract}

The automatic synthesis of policies for robotic-control tasks through reinforcement learning relies on a reward signal that simultaneously captures many possibly conflicting requirements. 
In this paper, we in\-tro\-duce a novel, hierarchical, potential-based reward-shaping approach (HPRS) for defining effective, multivariate rewards for a large family of such control tasks. We formalize a task as a partially-ordered set of safety, target, and comfort requirements, and define an automated methodology to enforce a natural order among requirements and shape the associated reward. Building upon potential-based reward shaping, we show that HPRS preserves policy optimality. 
Our experimental evaluation demonstrates HPRS's superior ability in capturing the intended behavior, resulting in task-satisfying policies with improved comfort, and converging to optimal behavior faster than other state-of-the-art approaches. 
We demonstrate the practical usability of HPRS on several robotics applications and the smooth sim2real transition on two autonomous-driving scenarios for F1TENTH race cars.
\end{abstract}

\section{Introduction}
Reinforcement learning (RL) is an increasingly popular method for training autonomous agents to solve complex tasks in sophisticated  environments~\cite{Nature/Mnih2015HumanLevelControlThroughDRL,arxiv/Lillicrap2015ContinuousControlWtDRL,Nature/Silver2017MasteringGameOfGoWithoutHumanKnowledge}. 
A key ingredient in RL is the reward function, a user-provided reinforcement signal, rewarding or penalizing the agent's behavior. 
Autonomous agents are becoming increasingly complex and are expected to satisfy numerous, potentially conflicting requirements. 
Since the reward function must capture all the desired aspects of the agent's behavior, a significant research effort has been invested in reward shaping over the past years~\cite{ICML/Ng19999PolicyInvarianceUnderRewardTransformation, ICML/Laud2003InfluenceOfRewardOnTheSpeedOfRL}. 

There are two major challenges in defining meaningful rewards, best illustrated with an autonomous-driving (AD) application. 
The first arises from mapping numerous requirements into a single scalar reward signal. 
In AD, there are more than $200$ rules that need to be considered when assessing the course of action ~\cite{DBLP:conf/icra/CensiSWYPFF19}. 
The second arises from the highly non-trivial task of determining the relative importance of the different requirements. 
In this realm, there are a plethora of regulations, ranging from safety and traffic rules, to performance, comfort, legal, and ethical requirements.

In order to address these challenges, we introduce HPRS, a novel, Hierarchical, Potential-based, Reward-Shaping technique to define the reward function from the formal requirements in a systematic fashion.
%
We use an expressive language to formalize safety, target, and comfort requirements, and consider a task as a partially-ordered set of requirements.

In contrast to classical potential-based approaches, we exploit the partial order and the quantitative evaluation of the individual requirements in the HPRS function. 
Unlike multi-objective approaches, HPRS defines only one multivariate multiplicative objective, which optimizes all the requirements simultaneously.
The potential formulation also allows us to provide theoretical guarantees on HPRS soundness~\cite{ICML/Ng19999PolicyInvarianceUnderRewardTransformation}. 
Finally, in contrast to logic-based approaches, which compute the reward on transition sequences~\cite{li2017tl-rewards, li2018tl-policy-search, IROS/Anand2019StructuredRewardShapingUsingSTL}, 
we provide a reward in every time-step. 
This way, HPRS avoids delaying reward computation over time and mitigates the temporal credit-assignment problem, where a deferred reward is not efficiently propagated to the preceding transitions.

Our approach builds on top of three major components:
\begin{itemize}
    \item An expressive formal specification language capturing 
    classes of requirements that often occur in control tasks.
    \item An additional specification layer, allowing to group sets of requirements and define priorities among them.
    \item An automatic procedure for generating a reward, following the order relation among the different requirements.
\end{itemize}

The advantage of our approach is the seamless passage from task specifications to learning optimal control policies that satisfy the associated requirements, while 
relieving the engineer from the burden of manually shaping rewards. 

We evaluated HPRS on three standard con\-ti\-nu\-ous-control benchmarks (lunar lander, bipedal walker classic and hardcore) and two autonomous driving scenarios (stand-alone and follow the leader). 
Our experimental results show that HPRS is very competitive compared to state-of-the-art approaches. 
Moreover, we deploy the resulting policies on F1TENTH race cars~\cite{okelly2020_f1tenth}, demonstrating the practical usability of HPRS in non-trivial real-world robotics systems.

\begin{figure*}[t]
    \centering
    \includegraphics[width=\textwidth]{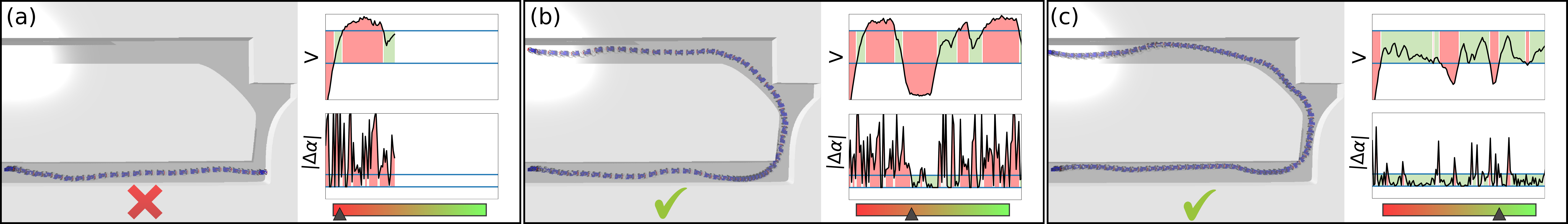}
    \vspace*{-1ex}
    \caption{
    The progress of a car driving on a track with one 90-degree turn. (a)~The car fails to be safe by crashing into the wall. (b)~The car safely progresses towards completing the lap, but in doing so, it fails to keep the velocity $V$ within the desired bounds and results in an unnecessary jerky steering $\Delta\alpha$. (c)~The car also maximizes the comfort requirements, as indicated by the increased number of green zones in the plots.
    }
    \label{fig:racecar_trajectories}
    \vspace*{-3ex}
\end{figure*}

\section{Motivating Example}

We motivate our work with an \textit{autonomous-driving task}: A \textit{car} drives around a track delimited by \textit{walls} by controlling its speed and steering angle.
We say the car completes a \textit{lap} when it drives around the track till its starting position.

The task has seven \textit{requirements}:
(1)~the car shall complete 1 lap in bounded time;
(2)~the car shall never collide against the walls;
(3)~the car shall drive in the center of the track;
(4)~the car shall keep a speed above a minimum value;
(5)~the car shall keep a speed below a maximum value;
(6)~the car shall drive with a comfortable steering angle;
(7)~the car shall send smooth control commands to the actuators;

A moment of thought reveals that these requirements might interfere with each other. 
For example, a car always driving above the minimum speed (Requirement 4), while steering below the maximum angle (Requirement 6), would have a limited-turn curvature. 
Any track layout containing a turn with a curvature larger than this limit would result in a collision, thus violating Requirement 2. 

Furthermore, if the policy is bang-bang, that is, it drives with 
high-frequency saturated actuation only, the resulting behavior is uncomfortable to passengers and not transferable to real hardware because of actuator limitations.
In Fi\-gure~\ref{fig:racecar_trajectories}, we show various intended and unintended behaviors.

In this example, it also becomes evident that some requi\-re\-ments must have precedence over others. 
We consider \textit{safety} as those requirements that fundamentally constrain the policy behavior,
such as a catastrophic collision against the walls.
Therefore, we interpret a safety violation as one compromising the validity of the entire episode.
Lap completion (Requirement 1) is also a unique requirement that represents 
the agent's main objective, or \textit{target}, and in essence, its whole reason to be. 
%
After the safety requirement, this comes next in the hierarchy of importance. 
Explicitly, it means that we are willing to sacrifice the rest of the requirements (Requirements 3-7) 
in order to complete a collision-free lap around the track. 
These requirements are, therefore, soft constraints that should be optimized as long as they do not interfere with safety and target. We
call them \textit{comfort}.


In summary, we pose the following research question in this paper: 
\emph{Is there a principled way to shape an effective reward that takes into account all the task requirements in the order of importance mentioned above?} In the rest of this paper, we will illustrate the necessary steps leading to a positive answer, on hand of the motivating example.

\section{Related Work}

Specifying reward functions for decision-making algorithms is a long-studied problem in the RL community. 
A poorly designed reward might not capture the actual objective and result in problematic or inefficient behaviors~\cite{arxiv/Amodei2016ConcreteProblemsAISafety}.
Therefore, shaping the reward helps to effectively steer the RL agent towards favorable behaviors~\cite{ICML/Ng19999PolicyInvarianceUnderRewardTransformation, ICML/Laud2003InfluenceOfRewardOnTheSpeedOfRL}.

\subsubsection{RL with Temporal Logic}
Much prior work adopts \emph{temporal logic} (TL) in RL.
Some of it focuses on the decomposition of a complex task into many sub-tasks~\cite {DBLP:journals/corr/Jothimurugan2021_CompositionalRLFromLogicalSpecs, icarte2018ltl-tasks}. Other formulations are tailored to tasks specified in TL~\cite{DBLP:conf/rss/FuTopcu2014_PACLearningWithTLConstraints, li2018tl-policy-search, Jiang2021TLRewardShaping, DBLP:conf/icml/Icarte2018_reward_machines}.
We consider the problem of reward shaping in the standard cumulative RL setting.
Several works exploit the quantitative semantics of TL (i.e., STL and its variants) to systematically derive a reward ~\cite{li2017tl-rewards,arxiv/Jones2015RobustSatOfTLSpecViaRL,IROS/Anand2019StructuredRewardShapingUsingSTL}.
However, they describe the task as a monolithic formula and compute the reward by either looking at the complete past
sequence~\cite{li2017tl-rewards}, or a sub-sequence~\cite{IROS/Anand2019StructuredRewardShapingUsingSTL}.
Thus, they construct sparse rewards and suffer from the credit-assignment problem. In contrast, we interpret a task as the composition of different requirements and provide a reward at every step. 
This approach is more in-line with cumulative RL formulations used in robotics and completely agnostic to the learning algorithm.

\subsubsection{Multi-Objective RL} 
Multi-Objective RL (MORL) studies the optimization of multiple and often conflicting objectives.
MORL algorithms learn single or multiple policies ~\cite{roijers2013mo-sequential,liu2015morl}. 
There exist several techniques to combine multiple reward signals into a single scalar value (i.e., scalarization), such as linear or non-linear projections ~\cite{natarajan2005multi-criteria,barrett2008multiple-criteria,vanMoffaert2013morl}.
Other approaches formulate structured rewards by imposing or assuming a preference ranking on the objectives and finding an equilibrium among them \cite{icml/GaborKS98MultiCriteriaRL,NIPS/Shelton2000BalancingMultipleSourcesOfRewardInRL,yun2010ranking,PMLRr/Abels19DynamicWeightsInMODRL}.
We focus on the single-policy setting and propose a multivariate multiplicative way to combining requirements~\cite{DBLP:books_RusselNorvig_AIAModernApproach}. We exploit the natural interpretation of the requirement classes to provide an unambiguous interpretation of task satisfaction, without the need to deal with \textit{Pareto-optimal} solutions.

A similar approach has been proposed in \cite{DBLP:conf/ijcnn/Brys2014_MultiObjectivizationOfRLProblems},
where the authors show that decomposing the task specification in many requirements can improve the learning process.
While these approaches still rely on the arbitrary choice of weights for each requirement,
we focus on defining a systematic methodology to produce a reward signal.
For completeness, in the experimental phase, we compare our approach to various instances of the linear-scalarization method adopted in \cite{DBLP:conf/ijcnn/Brys2014_MultiObjectivizationOfRLProblems}, and show the negative impact of having an arbitrary choice of static weights.


\subsubsection{Hierarchically Structured Requirements.}
Partially ordering requirements has been proposed before but in different settings.
The \textit{rulebook} formalism uses a set of prioritized requirements for evaluating behaviors produced by a planner~\cite{DBLP:conf/icra/CensiSWYPFF19}, 
or generating adversarial tests~\cite{DBLP:conf/rv/ViswanadhaRV2021_MultiObjectiveFalsificationScenicVerifAI}.
The complementary inverse RL approach in~\cite{journals/ral/Puranic2021LerarningFromDemonstrationUsingSTLInStochAndContDomains} learns dependencies among formal requirements from demonstrations.
However, while they learn dependencies from data, we infer them from requirement classes and use them in reward shaping.

\section{Main Contribution}
In this section, we present our main contribution: 
\emph{A method for automatically generating a reward-shaping function from a plant definition and a set of safety, target, and comfort requirements}.
%
In order to make this method accessible, we first introduce a \textit{formal language} allowing to formulate the requirements mentioned above. Our method then:
\begin{itemize}
    \item \textit{Step~1:} Infers the priority among requirements and formulates a task as a partially-ordered set of requirements.
    \item \textit{Step~2:} Extends the plant to an MDP by adding a sparse-reward signal and the episode-termination conditions.
    \item \textit{Step~3:} Extends the reward with a continuous HPRS by hierarchically evaluating the individual requirements.
\end{itemize}

\subsection{Requirements-Specification Language}

We formally define a set of expressive operators to capture requirements often occurring in continuous-control problems.
Considering atomic predicates $p \doteq f(s)\,{\geq}\,0$ over observable states $s\,{\in}\,S$, 
we extend existing task-specification languages 
(e.g., SpectRL \cite{DBLP:journals/corr/Jothimurugan2021_CompositionalRLFromLogicalSpecs})
and define requirements as follows:
\begin{align}
\label{task:syntax}
\begin{split}
\varphi  \doteq  &\ \texttt{achieve}\ p  ~|~ \texttt{conquer}\ p \\
                 &\  \texttt{ensure}\ p  \,\,\,\,\,\,|~ \texttt{encourage}\ p 
\end{split}
\end{align}
Commonly, a task can be defined as a set of requirements from three basic classes: {\em safety}, {\em target}, and {\em comfort}. Safety requirements, of the form $\texttt{ensure}\ p$, are naturally associated to an invariant condition $p$. Target requirements, of the form  $\texttt{achieve}\ p$ or  $\texttt{conquer}\ p$, formalize the one-time or respectively the persistent achievement of a goal within an episode. Finally, comfort requirements, of the form $\texttt{encourage}~p$, introduce the soft satisfaction of $p$, as often as possible, without compromising task satisfaction.

Let $\tau = (s_0, a_1, s_1, a_2, \ldots)$ denote an episode of $|\tau|=t$ steps, and let $\Tau$ be the set of all such traces. 
Each requirement $\varphi$ induces a Boolean function $\sigma:\,\Tau\,{\rightarrow}\,\BB$ evaluating whether an episode $\tau\,{\in}\,\Tau$ satisfies the requirement $\varphi$. We define the requirement satisfaction function $\sigma$ as follows:
\begin{align*}
 &\sigma(\texttt{achieve}\ p, \tau) &    & \text{ iff } \exists i \leq t \text{ s.t. }f(s_i)\geq0  \\
 &\sigma(\texttt{conquer}\ p, \tau) &    & \text{ iff }    \exists i \leq t \text{ s.t. } \forall j \geq i, f(s_j)\geq0  \\
 &\sigma(\texttt{ensure}\ p, \tau)  &    &  \text{ iff } \forall i \leq t \text{ s.t. } f(s_i) \geq 0 \\
 &\sigma(\texttt{encourage}\ p, \tau) &    &\text{ iff } \textsf{true}                           
\end{align*}

%

\begin{example}
\label{ex:task}

Consider the motivating example, and let us formally specify its requirements. 
The state $s = (x, y, \theta, v, \dot{\theta})$ 
consists of $x, y, \theta$ for the car position and heading in global coordinates,
$v$ and $\dot{\theta}$ are the car speed and rotational velocities, respectively.
The control action is $a = (\nu, \alpha)$ where 
$\nu$ denotes the desired speed, and $\alpha$ the steering angle.

We first define: 
(1) $L: S \rightarrow [0, 1]$ a lap progress function which maps the car position to the fraction of track that has been driven from the starting position;
(2) $d_{\emph{walls}}: S \rightarrow \RR$ a distance function which returns the distance of the car to the closest wall;  
(3) $d_{\emph{center}}: S \rightarrow \RR$ a distance function which returns the distance of the car to the centerline;
(4) the maximum deviation from the centerline $d_{\emph{comf}}$ that we consider tolerable;
(5) the maximum steering angle $\alpha_{\emph{comf}}$ that we consider being comfortable to drive straight;
(6) the minimum and maximum speed $v_{min}, v_{max}$ that define the speed limits;
(7) the maximum tolerable change in controls $\Delta a$ that we consider to be comfortable;
Then, the task can be formalized with the requirements
reported in Table~\ref{tab:requirements}.

\begin{table}[t]
    \caption{Driving example -- formalized requirements}
    \label{tab:requirements}
    \centering
    \begin{tabular}{|l|l|l|l|}
    \hline
    Req Id & Formula Id & Formula \\
    \hline \hline
    Req1 & $\varphi_1$ & $\texttt{achieve } L(s) = 1.0$ \\
    Req2 & $\varphi_2$ & $\texttt{ensure } d_{\emph{walls}}(s) > 0$ \\
    Req3 & $\varphi_3$ & $\texttt{encourage } d_{\emph{center}}(s) \leq d_{\emph{comf}}$ \\
    Req4 & $\varphi_4$ & $\texttt{encourage } v \geq v_{min}$ \\
    Req5 & $\varphi_5$ & $\texttt{encourage } v \leq v_{max}$ \\
    Req6 & $\varphi_6$ & $\texttt{encourage } |\alpha| \leq \alpha_{\emph{comf}}$ \\
    Req7 & $\varphi_7$ & $\texttt{encourage } |a| \leq \Delta a$ \\
    \hline
    \end{tabular}
\vspace*{-4ex}
\end{table}
\end{example}

\subsection{ A Task as a Partially-Ordered Set of Requirements}

We formalize a task by a partially-ordered set of formal requirements $\Phi$, assuming that the target is unique and unambiguous. Formally, $\Phi = \Phi_S \uplus \Phi_T \uplus \Phi_C$ such that:
$$
\begin{array}{lc}
    \Phi_S := \{ \varphi \, | \,  \varphi \doteq \texttt{ensure}\ p \} \\
    \Phi_C := \{ \varphi \, | \,  \varphi \doteq \texttt{encourage}\ p \} \\
    \Phi_T := \{ \varphi \, | \, \varphi \doteq \texttt{achieve}\ p \,\vee\, \varphi \doteq \texttt{conquer}\ p \} 
\end{array}
$$
\noindent The target requirement is required to be unique ($|\Phi_T| = 1$). 



We use a very natural interpretation of importance among the class of requirements, which considers decreasing importance from safety, to target, and to comfort requirements.

Formally, this natural interpretation of importance defines a (strict) partial order relation $\strictorder$ on $\Phi$ as follows:
$$
\varphi \strictorder \varphi' \text{ iff } \left( \varphi \in \Phi_S \wedge \varphi' \not \in \Phi_S\right) \vee \left( \varphi \in \Phi_T \wedge \varphi' \in \Phi_C \right)
$$

%

The resulting pair $(\Phi, \strictorder)$ forms a partially-ordered set of requirements and defines our task. 
Extending the satisfaction semantics to a set, we consider a task accomplished when all of its requirements are satisfied:
\begin{equation}
\label{eq:task_sat}
\sigma(\Phi, \tau) \text{ iff } \forall \varphi \in \Phi, \, \sigma(\varphi, \tau).
\end{equation}

\subsection{MDP Formalization of a Task}
\label{sec:mdp}

We assume that the plant (environment controlled by an autonomous agent) is given as $E\,{=}\,(S, S_0, A, P)$, where $S$ is the set of states, $S_0$ the set of initial states, $A$ is the set of actions, and $P(s'|s,a)$ is its dynamics, that is, the probability of reaching state $s'$ by performing action $a$ in state $s$.

Given an episodic task $(\Phi, \strictorder)$ over a bounded time horizon $T$, our goal is to automatically extend the environment $E$ to a Markov Decision Process (MDP) $M\,{=}\,(S,S_0,A,P,R)$. To this end, we define $R(s,a,s')$, the reward associated to the transition from state $s$ to $s'$ under action $a$, to satisfy $(\Phi, \strictorder)$.

\subsubsection{Episodes}
An episode ends when its task satisfaction is decided: either through a safety violation, timeout, or goal achievement. 
The goal-achievement evaluation depends on the target operator adopted: 
for $\texttt{achieve}\ p$ the goal is achieved when visiting at time $t\,{\leq}\,T$ a state $s_t$ that satisfies $p$; 
for $\texttt{conquer}\ p$ the goal is achieved if there is a time $i\,{\leq}\,T$ such that $p$ is satisfied for all $s_i,~i\,{\leq}\,t\,{\leq}\,T$.

\subsubsection{Base reward}
Given task $(\Phi, \strictorder)$, we first define a sparse reward incentivizing goal achievement. 
Let the property of the unique target requirement be $p \doteq f(s) \geq 0$. 
Then:
\begin{align*}
& R(s, a, s') =
\left\{ 
  \begin{array}{ c l }
    1           & \quad \textrm{if } f(s')\geq 0 \\
    0           & \quad \textrm{otherwise}
  \end{array}
\right.
\end{align*}

The rationale behind this choice is that we aim to teach the policy 
to reach the target and stay there as often as possible.
For $\texttt{achieve}\ p$, $R$ maximizes the probability of satisfying $p$. For $\texttt{conquer}\ p$, there is an added incentive to reach the target as soon as possible, and stay there until $T$.

The associated MDP is, in principle, solvable with any RL algorithm. However, while the sparse base reward $R$ can help solving simple tasks, where the target is easily achieved, it is completely ineffective in more complex control tasks.

\subsection{Hierarchical Potential-based Reward Shaping}
The main contribution is HPRS, our hierarchical potential-based reward shaping. 
This signal continuously provides feedback, guiding the agent towards the task satisfaction.
%

We assume the predicates $p(s)\,{\doteq}\,f(s)\,{\geq}\,0$ to be not trivially satisfied in all the states $s$,
otherwise they can be omitted by the specification.
Each signal is then bounded in $[l, u]$, for $l < 0 < u$.
We also define the negatively saturated signal $f_{-}(s)\,{=}\,min(0,f(s))$, 
and the following two signals:
$$
c(p, s) \doteq 1 - \frac{f_{-}(s)}{l}, \; 
b(p, s) \doteq \ONE_{\geq0}(f(s))
$$
where $\ONE_{\geq0}(\cdot)$ is an indicator function of non-negative numbers.
Both $c$ and $b$ are bounded in $[0, 1]$ where $1$ denotes the satisfaction of $p$ and $0$ its largest violation.
However, while $c$ is a continuous signal, $b$ is discrete, with values in $\{0, 1\}$. 

Using the signals $c$ and $b$, we now define the individual score $r$ for each requirement $\varphi\,{\in}\,\Phi$ as follows:
$$
r(\varphi, s) =
\left\{ 
\begin{array}{ll}
b(p, s) & \text{if}\ \varphi \in \Phi_S \\
c(p, s) & \text{otherwise}
\end{array}
\right.
$$

\begin{definition}
Let $(\Phi, \strictorder)$ be a task specification. 
Then the hierarchical potential function is defined as:
\begin{align}
\label{eq:hrs_pot}
\Psi(s) = \sum_{\varphi \in \Phi}  \left( \prod_{\varphi' : \varphi' \strictorder \varphi} r(\varphi', s) \right) \cdot r(\varphi, s)
\end{align}
\end{definition}

\vspace*{1mm}This potential function is a weighted sum over all requirements scores $r(\varphi,s)$. The weight of $r(\varphi,s)$ is the product of the scores $r(\varphi',s)$ of all the requirements $\varphi'$ that are strictly more important (hierarchically) than $\varphi$. 

The potential is thus a \textit{multivariate signal that combines the scores with multiplicative terms}~\cite{DBLP:books_RusselNorvig_AIAModernApproach},
according to the ordering defined in the task $(\Phi, \strictorder)$. A linear combination of scores, as typical in multi-objective scalarization, would assume independence among objectives and would not be expressive enough to capture their interdependence~\cite{DBLP:books_RusselNorvig_AIAModernApproach}. Crucially, the weights dynamically adapt at every step, too, according to the satisfaction degree of the requirements.

\begin{corollary}
The optimal policy for the MDP $M'$, 
where its reward $R'$ is defined with HPRS as:
\begin{equation}
    \label{eq:hrs}
    R'(s, a, s') = R(s, a, s') + \Psi(s') - \Psi(s)
\end{equation}
is also an optimal policy for the MDP $M$ with reward $R$.
\end{corollary}

This corollary shows that HPRS is preserving the policy optimality
for the considered undiscounted episodic setting.
It follows by the fact that $\Psi : S\,{\rightarrow}\,\RR$ is a potential function (depends only on the current state),
and by the results in \cite{ICML/Ng19999PolicyInvarianceUnderRewardTransformation}.

\subsection{Policy Assessment Metric}
\label{sec:policy_assessment_metric}

Since each reward formulation has its own scale, 
comparing the learning curves needs an external, unbiased assessment metric. To this end, we introduce a \textit{policy-assessment metric} (PAM) $F$, 
capturing the logical satisfaction of various requirements. 
We use the PAM to monitor the learning process and 
compare HPRS to state-of-the-art approaches.

Let $\Phi = \Phi_{S} \uplus \Phi_{T} \uplus \Phi_{C}$ be the set of requirements defining the task. 
Then, we define $F$ as follows:
$$
F(\Phi, \tau) =  
\sigma(\Phi_S, \tau) + 
\tfrac{1}{2} \sigma(\Phi_T, \tau) + 
\tfrac{1}{4}\sigma_{avg}(\Phi_C, \tau)
$$

\noindent where $\sigma(\Phi, \tau)\,{\in}\,\{0,1\}$ is the satisfaction function evaluated over $\Phi$ and $\tau$. 
We also define a time-averaged version for any comfort requirement $\varphi = \texttt{encourage } f(s) \geq 0$, as: 
$$
\sigma_{avg}(\varphi, \tau) = \sum_{i=1}^{|\tau|} \frac{\ONE_{\geq0}(f(s_i) )}{|\tau|} .
$$
\noindent{}Its set-wise extension computes the set-based average.

\begin{lemma}
Consider a task $(\Phi, \strictorder)$ and an episode $\tau$. Then, the following relations hold for $F$:\vspace*{-1mm}
\[\begin{array}{l}
F(\Phi, \tau) \geq 1.0 \leftrightarrow \sigma( \Phi_S, \tau),\quad 
F(\Phi, \tau) \geq 1.5 \leftrightarrow  \sigma(\Phi, \tau)
\end{array}\]
\end{lemma}

\vspace*{1mm}The proof follows from the construction of the PAM and the semantics of the task satisfaction defined in Equation~\ref{eq:task_sat}.

\section{Experimental Results}
\subsection{Experimental Setup}
To evaluate HPRS, we employ a state-of-the-art implementation of the SAC algorithm~\cite{haarnoja2018soft,stable-baselines3} on five use cases: 
a custom driving task with single and multiple cars, respectively, 
the lunar-lander with an obstacle, and the bipedal-walker both in the classic and hardcore versions. 
In each use case, we formalize a set of requirements $\Phi = \Phi_{S} \uplus \Phi_{T} \uplus \Phi_{P}$ and derive their partially-ordered set formulation $(\Phi, \strictorder)$.

\subsubsection{Use Cases}

The \emph{safe driving} task has already been presented as a motivating example. 
The \emph{follow leading-vehicle} task consists of the extension
with a non-controllable leading vehicle which the car aims to safely follow,
keeping a comfortable distance to it.
The agent does not access the full state but only the most-recent observations from LiDAR, noisy velocity estimates, and previous controls.
The safety requirements are extended to consider the collision with the leading vehicle, and the comfort requirements consider the control requirements 
and encourage the car to keep a comfortable distance, 
without any constraints on the car speed. 
We formulate two safety, one target, and four comfort requirements.
The \emph{lunar-lander} agent's objective 
is to land at the pad with coordinates $(0,0)$. In this example, we assume infinite fuel. Landing outside of the pad is also 
possible. We allow continuous actions to control the lander and 
add an obstacle to the environment in the vicinity of the landing pad, which makes the landing task harder. 
We formulate two safety, one target, and two comfort requirements. The \emph{bipedal-walker} robot's objective is to move forward towards the end of the field without falling. We consider two variants of this case study: the classical one with the flat terrain; and the hardcore one with holes and obstacles. We formulate one safety, one target, and four comfort requirements.

The formal specification of the safety, target, and comfort requirements in all use cases discussed above will be made available as an appendix of the full version of this paper.


\subsubsection{Reward Baselines}
We implemented HPRS as in Equation~\ref{eq:hrs}.
We compared it with the original reward formulation, defined by experts in each environment, indicated as \emph{Shaped}, 
and three additional baselines from state-of-the-art work: 

\begin{itemize}
    \item \emph{TLTL}~\cite{li2017tl-rewards} specifies tasks in a bounded (Truncated) LTL variant, equipped with an infinity-norm quantitative semantics \cite{nickovic2004stl}.
    The quantitative evaluation of the episode is used as a reward. We employ the RTAMT monitoring tool to compute the episode robustness \cite{DBLP:conf/atva/Nickovic2020_RTAMT}.
    
    \item \emph{BHNR}~\cite{IROS/Anand2019StructuredRewardShapingUsingSTL} specifies tasks in a fragment of STL with the filtering semantics of \cite{DBLP:conf/hybrid/Rodionova2016_TemporalLogicAsFiltering}. The reward uses a sliding-window approach to produce more frequent feedback to the agent: at each step, it uses the quantitative semantics to evaluate a sequence of $H$ states.
    
    \item \emph{MORL}~\cite{DBLP:conf/ijcnn/Brys2014_MultiObjectivizationOfRLProblems} implements the multi-objectivization of the task and solves the multi-objective problem by linear scalarization. To assess the sensitivity to the choice of weights, we consider two variants: uniform weights \emph{MORL(unif.)}; and decreasing weights \emph{MORL(decr.)} where safety is more important than target, and target is more important than comfort.
\end{itemize}

\subsection{Experimental Evaluation}

\subsubsection{Comparison to baselines}
We compare \emph{HPRS} to the above baselines and 
empirically show its superior performance in properly capturing the desired requirements. We use PAM $F$ for a sound and unbiased comparison.

$F$ allows to categorize each episode $\tau$ as: 
(1)~satisfying safety, if $F(\Phi, \tau)\,{\geq}\,1$,
(2)~satisfying safety and target, if $F(\Phi, \tau)\,{\geq}\,1.5$, and 
(3)~additionally maximizing comfort, if $F(\Phi, \tau)$ is close to $1.75$. We emphasize that $F$ is not used for training. Hence, it should not be used to evaluate the convergence of the RL algorithm in the training process.

\begin{figure*}
    \centering
    \includegraphics[width=0.9\textwidth]{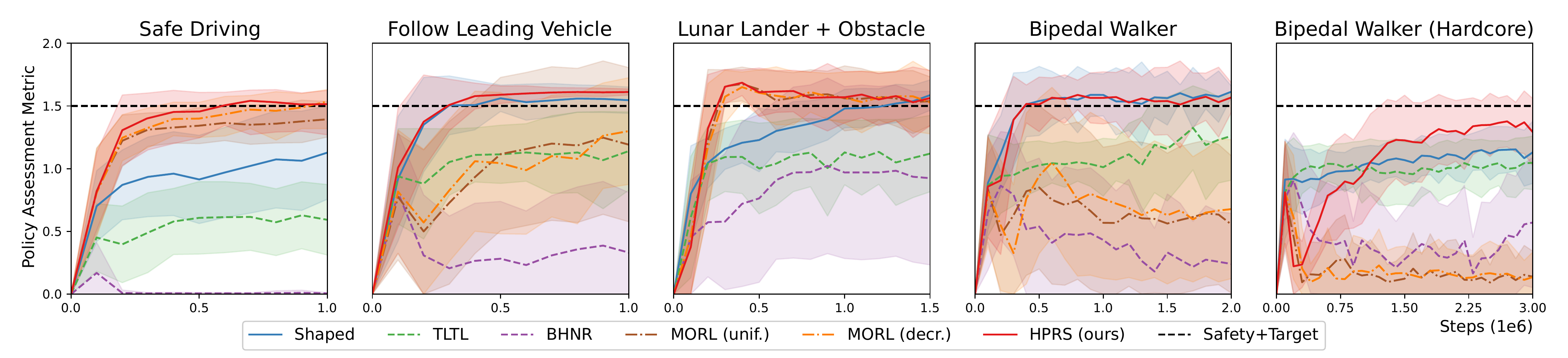}
    \vspace*{-1ex}
    \caption{Performance with respect to PAM $F$ (the policy-assessment metric) defined in \ref{sec:policy_assessment_metric}. 
    This metric has soundness guarantees in the evaluation of the task specification and is used to compare agents trained with different rewards. 
    The threshold, drawn as a black dashed horizontal line, indicates the task satisfaction. All performance functions report the mean (as a solid curve) and the standard deviation (as a colored shadow) over 10 seeds.
    }
    \label{fig:learning_curves}
    \vspace*{-3ex}
\end{figure*}

Figure \ref{fig:learning_curves} shows that HPRS has superior performance and faster convergence to task-satisfying policies, 
even better than the shaped reward in most of the tasks. 
The other approaches are not competitive to learn a policy for tasks with a high number of requirements. 

\subsubsection{Offline evaluation of the learned behaviors}

Despite the definition of a custom metric, capturing complex behaviors with a single scalar remains challenging. For this reason, we perform an extensive offline evaluation by comparing the policies (agents) trained with HPRS against the ones trained by using the other baseline rewards.
We provide evidence of the emergent behaviors in the submitted video.

We evaluate each trained policy with respect to each class of requirements in $50$ random episodes.
Table \ref{table:offline_evaluation} reports the 
Success Rate for incremental sets of Safety (S), Safety and Target (S+T), 
Safety, Target, and Comfort (S+T+C). 

\begin{table}[ht]
\centering
\resizebox{\columnwidth}{!}{
\begin{tabular}{ll|ccc}
    \textbf{Environment}    & \textbf{Reward}       & \textbf{S} & \textbf{S+T}        & \textbf{S+T+C} \\
                            &                       & Succ.Rate ($\%$) & Succ.Rate ($\%$)   & Succ.Rate ($\%$)     \\ \hline
    
    Safe Driving            & Shaped                & $0.74$ & $0.74$ & $0.20$ \\
    Safe Driving            & TLTL                  & $0.38$ & $0.32$ & $0.10$ \\
    Safe Driving            & BHNR                  & $0.00$ & $0.00$ & $0.00$ \\
    Safe Driving            & MORL(unif.)           & $0.99$ & $0.69$ & \textbf{0.32} \\
    Safe Driving            & MORL(decr.)           & $0.96$ & $0.75$ & \textbf{0.35} \\
    Safe Driving            & \textbf{HPRS(ours)}   & $0.97$ & $0.73$ & \textbf{0.33} \\
    \hline
    Follow Leading Vehicle   & Shaped                & $0.97$ & $0.97$ & $0.34$ \\
    Follow Leading Vehicle   & TLTL                  & $0.94$ & $0.12$ & $0.06$ \\
    Follow Leading Vehicle   & BHNR                  & $0.31$ & $0.00$ & $0.00$ \\
    Follow Leading Vehicle   & MORL(unif.)           & $0.74$ & $0.73$ & $0.35$ \\
    Follow Leading Vehicle   & MORL(decr.)           & $0.82$ & $0.81$ & $0.37$ \\
    Follow Leading Vehicle   & \textbf{HPRS(ours)}   & $1.00$ & $0.99$ & \textbf{0.46} \\ 
    \hline
    Lunar Lander            & Shaped                & $0.98$        & $0.72$ & $0.72$        \\
    Lunar Lander            & TLTL                  & $0.92$        & $0.00$ & $0.00$        \\
    Lunar Lander            & BHNR                  & $0.51$        & $0.49$ & $0.49$        \\
    Lunar Lander            & MORL(unif.)           & $0.91$        & $0.91$ & \textbf{0.90} \\
    Lunar Lander            & MORL(decr.)           & $0.94$        & $0.91$ & \textbf{0.91} \\
    Lunar Lander            & \textbf{HPRS(ours)}   & $0.91$        & $0.91$ & \textbf{0.89} \\ 
    \hline
    Bipedal Walker          & Shaped                & $0.99$        & $0.99$        & \textbf{0.51} \\
    Bipedal Walker          & TLTL                  & $0.96$        & $0.45$        & $0.27$        \\
    Bipedal Walker          & BHNR                  & $0.21$        & $0.00$        & $0.00$        \\
    Bipedal Walker          & MORL(unif.)           & $0.40$        & $0.40$        & $0.19$        \\
    Bipedal Walker          & MORL(decr.)           & $0.43$        & $0.43$        & $0.20$        \\
    Bipedal Walker          & \textbf{HPRS(ours)}   & $0.96$        & $0.96$        & \textbf{0.48}        \\ 
    \hline
    Bipedal Walker (Hardcore)   & Shaped                & $0.84$        & $0.29$        & $0.17$    \\
    Bipedal Walker (Hardcore)   & TLTL                  & $0.98$        & $0.00$        & $0.00$    \\
    Bipedal Walker (Hardcore)   & BHNR                  & $0.55$        & $0.00$        & $0.00$    \\
    Bipedal Walker (Hardcore)   & MORL(unif.)           & $0.07$        & $0.03$        & $0.02$    \\
    Bipedal Walker (Hardcore)   & MORL(decr.)           & $0.06$        & $0.03$        & $0.02$    \\
    Bipedal Walker (Hardcore)   & \textbf{HPRS(ours)}   & $0.85$        & $0.85$        & \textbf{0.44} \\ 
    \hline
    
\end{tabular}}
\caption{
Evaluation of trained agents over classes of requirements: Safety (S), Target (T), Comfort (C). 
Results $<5\%$ close to the best-performing reward shaping are marked in bold.
}
\vspace*{-4ex}
\label{table:offline_evaluation}
\end{table}

Policies learned with HPRS consistently complete the task in most evaluations, competing with hand-crafted rewards and proving their ability in trading-off the different requirements. 
While other baselines struggle in capturing the correct objective and do not show consistent performance across different domains,
we highlight that HPRS is $<5\%$ close to the best-performing approach in all the tasks.

Logic-based approaches, such as \emph{TLTL} and \emph{BHNR}, 
consider the task as a unique specification and result in policies that either eagerly maximize the progress towards the target, 
resulting in unsafe behaviors, or converge to an over-conservative behavior that never achieves task completion.
This observation highlights the weakness of these approaches 
when dealing with many requirements, 
because the dominant requirement could mask out the others, 
even if normalized adequately to the signal domain.

Multi-objective approaches are confirmed to be sensitive to weights selection. 
While their performance is competitive in some of the tasks, 
they perform poorly in more complex ones, such as those presented in the bipedal walker.

Finally, the Shaped reward results in policies capturing the desired behavior, 
confirming the good reward shaping proposed in the original environments. 
However, considering the current training budget, HPRS produces a more effective learning signal, 
resulting in better-performing policies.

\subsubsection{Ablation Study on Comfort Requirements}

We evaluate the impact of individual requirements in the hierarchical structure of HPRS. 
We focus on the comfort requirements that have the least priority and thus the minor influence on the value of the final reward. 
Specifically, we study how the comfort requirements improve the observed comfort. 
We set up an ablation experiment on them and compare the performance of the resulting policies. 

Table~\ref{table:eval_comfort} reports the evaluation 
for policies trained with (+Comfort), and without (-Comfort) comfort requirements. 
For each seed, we collect $50$ episodes 
and compute the ratio of satisfaction of comfort requirements over each episode.

\begin{table}[ht]
\centering
\begin{tabular}{l|cc}
                                    & \textbf{+Comfort}     & \textbf{-Comfort}       \\
    \textbf{Safe Driving}   & $\sigma_{\text{avg}}$ & $\sigma_{\text{avg}}$   \\ \hline
    Keep the center                 & $0.39 \pm 0.12$     & $0.33 \pm 0.10$         \\ 
    Min Velocity                    & $0.48 \pm 0.21$     & $0.89 \pm 0.06$         \\ 
    Max Velocity                    & $0.99 \pm 0.01$     & $0.36 \pm 0.14$         \\ 
    Comfortable steering            & $0.27 \pm 0.08$     & $0.08 \pm 0.04$         \\ 
    Smooth control                  & $0.70 \pm 0.07$     & $0.32 \pm 0.07$         \\ \hline
    \textbf{Follow Leading Vehicle} &                     &                       \\ \hline
    Min Distance                    & $0.55 \pm 0.22$     & $0.85 \pm 0.16$       \\ 
    Max Distance                    & $0.90 \pm 0.08$     & $0.61 \pm 0.18$       \\
    Comfortable steering            & $0.23 \pm 0.05$     & $0.15 \pm 0.04$       \\ 
    Smooth control                  & $0.78 \pm 0.09$     & $0.41 \pm 0.11$       \\ \hline
    \textbf{Lunar Lander}           &                       &                       \\ \hline
    Hull angle                      & $0.99 \pm 0.05$     & $0.99 \pm 0.02$       \\
    Hull angular velocity           & $0.97 \pm 0.07$     & $0.98 \pm 0.05$       \\ \hline
    \textbf{Bipedal Walker}         &                     &                       \\ \hline
    Hull angle                      & $0.80 \pm 0.17$     & $0.33 \pm 0.27$       \\
    Hull angular velocity           & $1.00 \pm 0.00$     & $0.99 \pm 0.01$       \\
    Vertical oscillation            & $0.98 \pm 0.01$     & $0.91 \pm 0.11$       \\
    Horizontal velocity             & $0.95 \pm 0.01$     & $0.92 \pm 0.03$       \\ \hline
    \textbf{Bipedal Walker Hardcore} &                    &                       \\ \hline
    Hull angle                      & $0.70 \pm 0.13$     & $0.29 \pm 0.14$       \\
    Hull angular velocity           & $1.00 \pm 0.00$     & $0.99 \pm 0.01$       \\
    Vertical oscillation            & $0.83 \pm 0.06$     & $0.75 \pm 0.09$       \\
    Horizontal velocity             & $0.94 \pm 0.05$     & $0.81 \pm 0.10$       \\ \hline
\end{tabular}
\caption{
Evaluation of policies trained with and without Comfort requirements ($\pm$Comfort). 
}
\label{table:eval_comfort}
\vspace*{-5ex}
\end{table}

In all the tasks, introducing comfort requirements positively impacts the evaluation.
While some of the requirements are almost always satisfied by both configurations, 
the satisfaction of other requirements significantly improves once comfort rules are introduced,
denoted by an increase of the mean satisfaction and a reduction in its standard deviation.
Especially in driving tasks, 
the smaller steering magnitude and smoother transition between consecutive controls
make the policy amenable to transfer to the real-world applications.

\section{Real-world Demonstration}

We validated the driving policies trained with HPRS in a real-world setting using the F1TENTH racing cars \cite{okelly2020_f1tenth}. 
The hardware platform consists of an off-the-shelf model race car chassis \emph{Traxxas Ford Fiesta ST}.
Actuation is provided by a brushless DC electric motor \emph{Traxxas Velineon 3351R} 
which is driven by a \emph{VESC 6 MkIV} electronic speed controller (ESC).
The distances to the walls are sensed by a \emph{Hokuyo UST-10LX} LiDAR sensor 
and the velocity estimate is directly read from the VESC.
The control loop is executed at a $10$ Hz rate 
on a \emph{NVIDIA Jetson Xavier NX} embedded computing platform.
We integrate the trained agent (sim2real) in a ROS node  within the F1TENTH software setup.
The speed and steering commands are passed to the auxiliary nodes, which automatically compute motor rpm and servo position.

\begin{figure}
    \centering
    \includegraphics[width=0.9\linewidth]{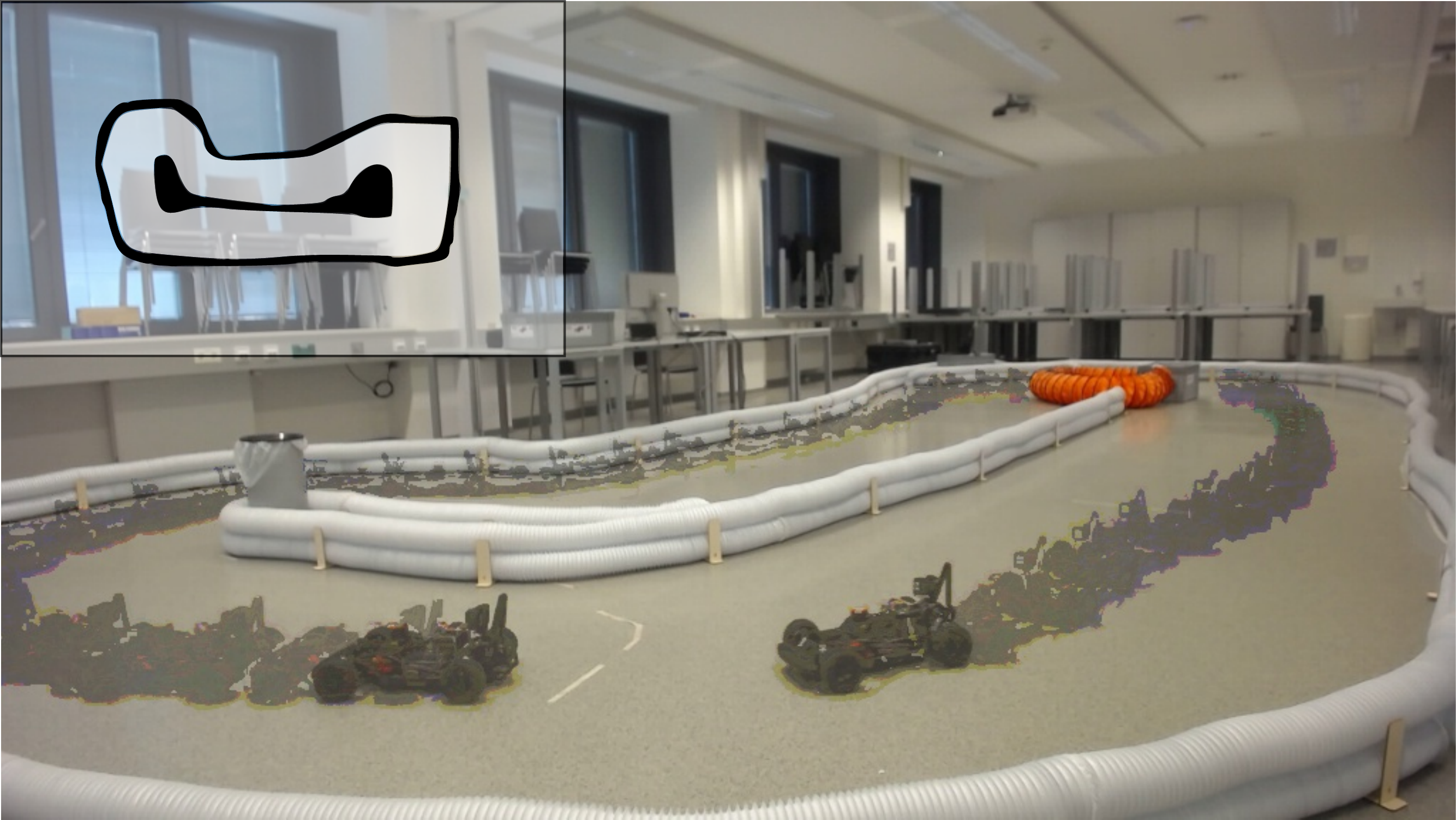}
    \caption{Demonstration of a successful deployment (sim2real) of our policy on an F1TENTH car. Inset image shows the track layout for this example.}
    \label{fig:real_car}
    \vspace*{-3ex}
\end{figure}

We train the policy in simulation \cite{10.1109/ICRA_brunnbauer_racing_dreamer}, 
sampling the car position and the simulation parameters (e.g., mass, sensor noise, actuator gains) to robustly transfer it to the real world \cite{tobin2017domainrandomization}.
Figure \ref{fig:real_car} shows a successful deployment for the Safe-Driving task.
The attached video shows: (1) the car smoothly driving along the track,
and (2) the car safely following a leading vehicle while keeping a comfortable distance.

\section{Conclusions}
This paper introduced HPRS, a novel, hierarchical, po\-ten\-tial-based reward-shaping method and tool for tasks ${(\Phi,\strictorder)}$, 
consisting of a partially-ordered set of safety, target, and comfort requirements. We showed that HPRS 
performs better than state-of-the-art approaches on five continuous-control benchmarks. We also demonstrated that HPRS facilitates a smooth sim2real transition on the two F1TENTH driving benchmarks.
The idea of automatically shaping rewards from specifications 
possessing an evaluation procedure, is general, and agnostic to the plant and the RL algorithm adopted. 
%

In subsequent work, we intend to consider more expressive operators and study the formalization of requirements beyond safety, progress, and comfort, 
such as  the ethical, legal, and performance objectives.
Both our code and the supplementary material are freely available at the following repository: \href{http://github.com/EdAlexAguilar/reward_shaping}{\url{github.com/EdAlexAguilar/reward_shaping}}.

\section*{Acknowledgement}
\noindent L.B. is supported by the Doctoral College Resilient Embedded Systems.
This work has received funding from the EU's Horizon 2020 research and innovation programme under grant No 956123 and from the Austrian FFG ICT of the Future program under grant No 880811.
We thank Axel Brunnbauer for contributing in the early stage of this work.

\bibliographystyle{IEEEtran}
\bibliography{IEEEabrv, refs.bib}

\begin{thebibliography}{10}
\providecommand{\url}[1]{#1}
\csname url@rmstyle\endcsname
\providecommand{\newblock}{\relax}
\providecommand{\bibinfo}[2]{#2}
\providecommand\BIBentrySTDinterwordspacing{\spaceskip=0pt\relax}
\providecommand\BIBentryALTinterwordstretchfactor{4}
\providecommand\BIBentryALTinterwordspacing{\spaceskip=\fontdimen2\font plus
\BIBentryALTinterwordstretchfactor\fontdimen3\font minus
  \fontdimen4\font\relax}
\providecommand\BIBforeignlanguage[2]{{%
\expandafter\ifx\csname l@#1\endcsname\relax
\typeout{** WARNING: IEEEtran.bst: No hyphenation pattern has been}%
\typeout{** loaded for the language `#1'. Using the pattern for}%
\typeout{** the default language instead.}%
\else
\language=\csname l@#1\endcsname
\fi
#2}}

\bibitem{Nature/Mnih2015HumanLevelControlThroughDRL}
V.~Mnih, K.~Kavukcuoglu, D.~Silver, A.~A. Rusu, J.~Veness, M.~G. Bellemare,
  A.~Graves, M.~Riedmiller, A.~K. Fidjeland, G.~Ostrovski, \emph{et~al.},
  ``Human-level control through deep reinforcement learning,'' \emph{nature},
  vol. 518, no. 7540, pp. 529--533, 2015.

\bibitem{arxiv/Lillicrap2015ContinuousControlWtDRL}
T.~P. Lillicrap, J.~J. Hunt, A.~Pritzel, N.~Heess, T.~Erez, Y.~Tassa,
  D.~Silver, and D.~Wierstra, ``Continuous control with deep reinforcement
  learning,'' \emph{arXiv preprint arXiv:1509.02971}, 2015.

\bibitem{Nature/Silver2017MasteringGameOfGoWithoutHumanKnowledge}
D.~Silver, J.~Schrittwieser, K.~Simonyan, I.~Antonoglou, A.~Huang, A.~Guez,
  T.~Hubert, L.~Baker, M.~Lai, A.~Bolton, \emph{et~al.}, ``Mastering the game
  of go without human knowledge,'' \emph{nature}, vol. 550, no. 7676, pp.
  354--359, 2017.

\bibitem{ICML/Ng19999PolicyInvarianceUnderRewardTransformation}
A.~Y. Ng, D.~Harada, and S.~Russell, ``Policy invariance under reward
  transformations: Theory and application to reward shaping,'' in \emph{In
  Proceedings of the Sixteenth International Conference on Machine
  Learning}.\hskip 1em plus 0.5em minus 0.4em\relax Morgan Kaufmann, 1999, pp.
  278--287.

\bibitem{ICML/Laud2003InfluenceOfRewardOnTheSpeedOfRL}
A.~Laud and G.~DeJong, ``The influence of reward on the speed of reinforcement
  learning: An analysis of shaping,'' in \emph{Proceedings of the 20th
  International Conference on Machine Learning (ICML-03)}, 2003, pp. 440--447.

\bibitem{DBLP:conf/icra/CensiSWYPFF19}
A.~Censi, K.~Slutsky, T.~Wongpiromsarn, D.~S. Yershov, S.~Pendleton, J.~G.~M.
  Fu, and E.~Frazzoli, ``Liability, ethics, and culture-aware behavior
  specification using rulebooks,'' in \emph{International Conference on
  Robotics and Automation, {ICRA} 2019, Montreal, QC, Canada, May 20-24, 2019},
  2019, pp. 8536--8542.

\bibitem{li2017tl-rewards}
X.~Li, C.-I. Vasile, and C.~Belta, ``Reinforcement learning with temporal logic
  rewards,'' in \emph{2017 IEEE/RSJ International Conference on Intelligent
  Robots and Systems (IROS)}, 2017, pp. 3834--3839.

\bibitem{li2018tl-policy-search}
X.~Li, Y.~Ma, and C.~Belta, ``A policy search method for temporal logic
  specified reinforcement learning tasks,'' \emph{2018 Annual American Control
  Conference (ACC)}, pp. 240--245, 2018.

\bibitem{IROS/Anand2019StructuredRewardShapingUsingSTL}
A.~Balakrishnan and J.~V. Deshmukh, ``Structured reward shaping using signal
  temporal logic specifications,'' in \emph{2019 IEEE/RSJ International
  Conference on Intelligent Robots and Systems (IROS)}, 2019, pp. 3481--3486.

\bibitem{okelly2020_f1tenth}
M.~O'Kelly, H.~Zheng, D.~Karthik, and R.~Mangharam, ``F1tenth: An open-source
  evaluation environment for continuous control and reinforcement learning,''
  \emph{Proceedings of Machine Learning Research}, vol. 123, 2020.

\bibitem{arxiv/Amodei2016ConcreteProblemsAISafety}
D.~Amodei, C.~Olah, J.~Steinhardt, P.~Christiano, J.~Schulman, and D.~Man{\'e},
  ``Concrete problems in ai safety,'' \emph{ArXiv}, vol. abs/1606.06565, 2016.

\bibitem{DBLP:journals/corr/Jothimurugan2021_CompositionalRLFromLogicalSpecs}
\BIBentryALTinterwordspacing
K.~Jothimurugan, S.~Bansal, O.~Bastani, and R.~Alur, ``Compositional
  reinforcement learning from logical specifications,'' \emph{CoRR}, vol.
  abs/2106.13906, 2021. [Online]. Available:
  \url{https://arxiv.org/abs/2106.13906}
\BIBentrySTDinterwordspacing

\bibitem{icarte2018ltl-tasks}
R.~Toro~Icarte, T.~Q. Klassen, R.~Valenzano, and S.~A. McIlraith, ``Teaching
  multiple tasks to an rl agent using ltl,'' in \emph{Proceedings of the 17th
  International Conference on Autonomous Agents and MultiAgent Systems}, 2018,
  pp. 452--461.

\bibitem{DBLP:conf/rss/FuTopcu2014_PACLearningWithTLConstraints}
\BIBentryALTinterwordspacing
J.~Fu and U.~Topcu, ``Probably approximately correct {MDP} learning and control
  with temporal logic constraints,'' in \emph{Robotics: Science and Systems X,
  University of California, Berkeley, USA, July 12-16, 2014}, D.~Fox, L.~E.
  Kavraki, and H.~Kurniawati, Eds., 2014. [Online]. Available:
  \url{http://www.roboticsproceedings.org/rss10/p39.html}
\BIBentrySTDinterwordspacing

\bibitem{Jiang2021TLRewardShaping}
\BIBentryALTinterwordspacing
Y.~Jiang, S.~Bharadwaj, B.~Wu, R.~Shah, U.~Topcu, and P.~Stone,
  ``Temporal-logic-based reward shaping for continuing reinforcement learning
  tasks,'' \emph{Proceedings of the AAAI Conference on Artificial
  Intelligence}, vol.~35, no.~9, pp. 7995--8003, May 2021. [Online]. Available:
  \url{https://ojs.aaai.org/index.php/AAAI/article/view/16975}
\BIBentrySTDinterwordspacing

\bibitem{DBLP:conf/icml/Icarte2018_reward_machines}
R.~T. Icarte, T.~Klassen, R.~Valenzano, and S.~McIlraith, ``Using reward
  machines for high-level task specification and decomposition in reinforcement
  learning,'' in \emph{International Conference on Machine Learning}.\hskip 1em
  plus 0.5em minus 0.4em\relax PMLR, 2018, pp. 2107--2116.

\bibitem{arxiv/Jones2015RobustSatOfTLSpecViaRL}
A.~Jones, D.~Aksaray, Z.~Kong, M.~Schwager, and C.~Belta, ``Robust satisfaction
  of temporal logic specifications via reinforcement learning,'' 2015.

\bibitem{roijers2013mo-sequential}
D.~M. Roijers, P.~Vamplew, S.~Whiteson, and R.~Dazeley, ``A survey of
  multi-objective sequential decision-making,'' \emph{J. Artif. Int. Res.},
  vol.~48, no.~1, p. 67–113, Oct. 2013.

\bibitem{liu2015morl}
C.~Liu, X.~Xu, and D.~Hu, ``Multiobjective reinforcement learning: A
  comprehensive overview,'' \emph{IEEE Transactions on Systems, Man, and
  Cybernetics: Systems}, vol.~45, no.~3, pp. 385--398, 2015.

\bibitem{natarajan2005multi-criteria}
S.~Natarajan and P.~Tadepalli, ``Dynamic preferences in multi-criteria
  reinforcement learning,'' in \emph{Proceedings of the 22nd international
  conference on Machine learning}, 2005, pp. 601--608.

\bibitem{barrett2008multiple-criteria}
L.~Barrett and S.~Narayanan, ``Learning all optimal policies with multiple
  criteria,'' in \emph{Proceedings of the 25th international conference on
  Machine learning}, 2008, pp. 41--47.

\bibitem{vanMoffaert2013morl}
K.~Van~Moffaert, M.~M. Drugan, and A.~Nowé, ``Scalarized multi-objective
  reinforcement learning: Novel design techniques,'' in \emph{2013 IEEE
  Symposium on Adaptive Dynamic Programming and Reinforcement Learning
  (ADPRL)}, 2013, pp. 191--199.

\bibitem{icml/GaborKS98MultiCriteriaRL}
Z.~G{\'{a}}bor, Z.~Kalm{\'{a}}r, and C.~Szepesv{\'{a}}ri, ``Multi-criteria
  reinforcement learning,'' in \emph{Proceedings of the Fifteenth International
  Conference on Machine Learning {(ICML} 1998), Madison, Wisconsin, USA, July
  24-27, 1998}, J.~W. Shavlik, Ed.\hskip 1em plus 0.5em minus 0.4em\relax
  Morgan Kaufmann, 1998, pp. 197--205.

\bibitem{NIPS/Shelton2000BalancingMultipleSourcesOfRewardInRL}
C.~Shelton, ``Balancing multiple sources of reward in reinforcement learning,''
  in \emph{Advances in Neural Information Processing Systems}, T.~Leen,
  T.~Dietterich, and V.~Tresp, Eds., vol.~13.\hskip 1em plus 0.5em minus
  0.4em\relax MIT Press, 2001.

\bibitem{yun2010ranking}
Y.~Zhao, Q.~Chen, and W.~Hu, ``Multi-objective reinforcement learning algorithm
  for mosdmp in unknown environment,'' in \emph{2010 8th World Congress on
  Intelligent Control and Automation}, 2010, pp. 3190--3194.

\bibitem{PMLRr/Abels19DynamicWeightsInMODRL}
A.~Abels, D.~Roijers, T.~Lenaerts, A.~Now{\'e}, and D.~Steckelmacher, ``Dynamic
  weights in multi-objective deep reinforcement learning,'' in
  \emph{International Conference on Machine Learning}.\hskip 1em plus 0.5em
  minus 0.4em\relax PMLR, 2019, pp. 11--20.

\bibitem{DBLP:books_RusselNorvig_AIAModernApproach}
\BIBentryALTinterwordspacing
S.~Russell and P.~Norvig, \emph{Artificial Intelligence: {A} Modern Approach
  (4th Edition)}.\hskip 1em plus 0.5em minus 0.4em\relax Pearson, 2020.
  [Online]. Available: \url{http://aima.cs.berkeley.edu/}
\BIBentrySTDinterwordspacing

\bibitem{DBLP:conf/ijcnn/Brys2014_MultiObjectivizationOfRLProblems}
T.~Brys, A.~Harutyunyan, P.~Vrancx, M.~E. Taylor, D.~Kudenko, and A.~Now{\'e},
  ``Multi-objectivization of reinforcement learning problems by reward
  shaping,'' in \emph{2014 international joint conference on neural networks
  (IJCNN)}.\hskip 1em plus 0.5em minus 0.4em\relax IEEE, 2014, pp. 2315--2322.

\bibitem{DBLP:conf/rv/ViswanadhaRV2021_MultiObjectiveFalsificationScenicVerifAI}
K.~Viswanadha, E.~Kim, F.~Indaheng, D.~J. Fremont, and S.~A. Seshia, ``Parallel
  and multi-objective falsification with scenic and verifai,'' in
  \emph{International Conference on Runtime Verification}.\hskip 1em plus 0.5em
  minus 0.4em\relax Springer, 2021, pp. 265--276.

\bibitem{journals/ral/Puranic2021LerarningFromDemonstrationUsingSTLInStochAndContDomains}
A.~G. Puranic, J.~V. Deshmukh, and S.~Nikolaidis, ``Learning from
  demonstrations using signal temporal logic in stochastic and continuous
  domains,'' \emph{IEEE Robotics and Automation Letters}, vol.~6, no.~4, pp.
  6250--6257, 2021.

\bibitem{haarnoja2018soft}
\BIBentryALTinterwordspacing
T.~Haarnoja, A.~Zhou, P.~Abbeel, and S.~Levine, ``Soft actor-critic: Off-policy
  maximum entropy deep reinforcement learning with a stochastic actor,'' in
  \emph{Proceedings of the 35th International Conference on Machine Learning},
  ser. Proceedings of Machine Learning Research, J.~Dy and A.~Krause, Eds.,
  vol.~80.\hskip 1em plus 0.5em minus 0.4em\relax PMLR, 10--15 Jul 2018, pp.
  1861--1870. [Online]. Available:
  \url{http://proceedings.mlr.press/v80/haarnoja18b.html}
\BIBentrySTDinterwordspacing

\bibitem{stable-baselines3}
A.~Raffin, A.~Hill, M.~Ernestus, A.~Gleave, A.~Kanervisto, and N.~Dormann,
  ``Stable baselines3,'' \url{https://github.com/DLR-RM/stable-baselines3},
  2019.

\bibitem{nickovic2004stl}
O.~Maler and D.~Nickovic, ``Monitoring temporal properties of continuous
  signals,'' in \emph{Formal Techniques, Modelling and Analysis of Timed and
  Fault-Tolerant Systems}, Y.~Lakhnech and S.~Yovine, Eds.\hskip 1em plus 0.5em
  minus 0.4em\relax Berlin, Heidelberg: Springer Berlin Heidelberg, 2004, pp.
  152--166.

\bibitem{DBLP:conf/atva/Nickovic2020_RTAMT}
D.~Ni{\v{c}}kovi{\'c} and T.~Yamaguchi, ``Rtamt: Online robustness monitors
  from stl,'' in \emph{International Symposium on Automated Technology for
  Verification and Analysis}.\hskip 1em plus 0.5em minus 0.4em\relax Springer,
  2020, pp. 564--571.

\bibitem{DBLP:conf/hybrid/Rodionova2016_TemporalLogicAsFiltering}
A.~Rodionova, E.~Bartocci, D.~Nickovic, and R.~Grosu, ``Temporal logic as
  filtering,'' in \emph{Proceedings of the 19th International Conference on
  Hybrid Systems: Computation and Control}, 2016, pp. 11--20.

\bibitem{10.1109/ICRA_brunnbauer_racing_dreamer}
\BIBentryALTinterwordspacing
A.~Brunnbauer, L.~Berducci, A.~Brandstaetter, M.~Lechner, R.~Hasani, D.~Rus,
  and R.~Grosu, ``Latent imagination facilitates zero-shot transfer in
  autonomous racing,'' in \emph{2022 International Conference on Robotics and
  Automation (ICRA)}.\hskip 1em plus 0.5em minus 0.4em\relax IEEE Press, 2022,
  p. 7513–7520. [Online]. Available:
  \url{https://doi.org/10.1109/ICRA46639.2022.9811650}
\BIBentrySTDinterwordspacing

\bibitem{tobin2017domainrandomization}
J.~Tobin, R.~Fong, A.~Ray, J.~Schneider, W.~Zaremba, and P.~Abbeel, ``Domain
  randomization for transferring deep neural networks from simulation to the
  real world,'' in \emph{2017 IEEE/RSJ international conference on intelligent
  robots and systems (IROS)}.\hskip 1em plus 0.5em minus 0.4em\relax IEEE,
  2017, pp. 23--30.

\end{thebibliography}


\onecolumn
\appendix

In this appendix we provide our implementation details for the experiments and the training/evaluation. This includes the requirements and formal specifications for the lunar lander and bipedal walker studies. We also include the details of the environment modifications we did to all environments (mainly lunar lander).

\section{Training and Evaluation Details}
Table \ref{tab:training_parameters} includes the details of the training parameters used. 
For training, we evaluate the progress of the policy every $10\, 000$ steps, and use $10$ episodes for the evaluation. 
As mentioned in the main text, we use the SAC implementation from \cite{stable-baselines3}.  
Table \ref{tab:training_parameters} also reports the algorithm hyper-parameters, omitting the ones we keep to default values.


\begin{table}[h]
\begin{tabular}{ |l c c c c c| }
\hline
 & Safe Driving & Follow Leading Vehicle & Lunar Lander + Obstacle & Bipedal Walker & Bipedal Walker (Hardcore)\\
\hline
\multicolumn{6}{|c|}{\textbf{Training and Evaluation Parameters}} \\
\hline
num\_steps & $1\text{e}6$ & $1\text{e}6$ & $1.5\text{e}6$ & $2\text{e}6$ & $3\text{e}6$   \\ \hline
evaluate\_every  & $1\text{e}4$ & $1\text{e}4$ & $1\text{e}4$ & $1\text{e}4$ & $1\text{e}4$  \\ \hline
num\_eval\_episodes & 10 & 10 & $10$ & $10$ & $10$\\ \hline
\multicolumn{6}{|c|}{\textbf{SAC Implementation Parameters}} \\ \hline
buffer\_size & $3\text{e}5$ & $3\text{e}5$ & $3\text{e}5$ & $1\text{e}6$ & $1\text{e}6$ \\ \hline
learning\_starts & $1\text{e}2$ & $1\text{e}2$ & $1\text{e}4$ & $1\text{e}2$ & $1\text{e}2$ \\ \hline
batch\_size & 256 & 256  & 256  & 256  & 256\\ \hline
tau & $0.005$ & $0.005$ & $0.01$ & $0.005$ & $0.005$\\ \hline
net\_architecture & DEF & Qf:$[256,256]$,   pi:$ [64,64]$ & $[400, 300]$ & DEF & DEF \\\hline
\end{tabular}
\begin{center}
\caption{Parameters used for training all environments. Omitted values are default (DEF) and common across all environments, e.g. $\gamma = 0.99$, learning\_rate$=0.0003$, gradient\_steps$=1$, target\_update\_interval$=1$, entropy\_coefficient$=$`auto'.}
\label{tab:training_parameters}
\end{center}
\end{table}

\section{Environment Descriptions}


\subsection{Safe Driving}
\begin{figure*}[ht]
    \centering
    \includegraphics[width=\linewidth]{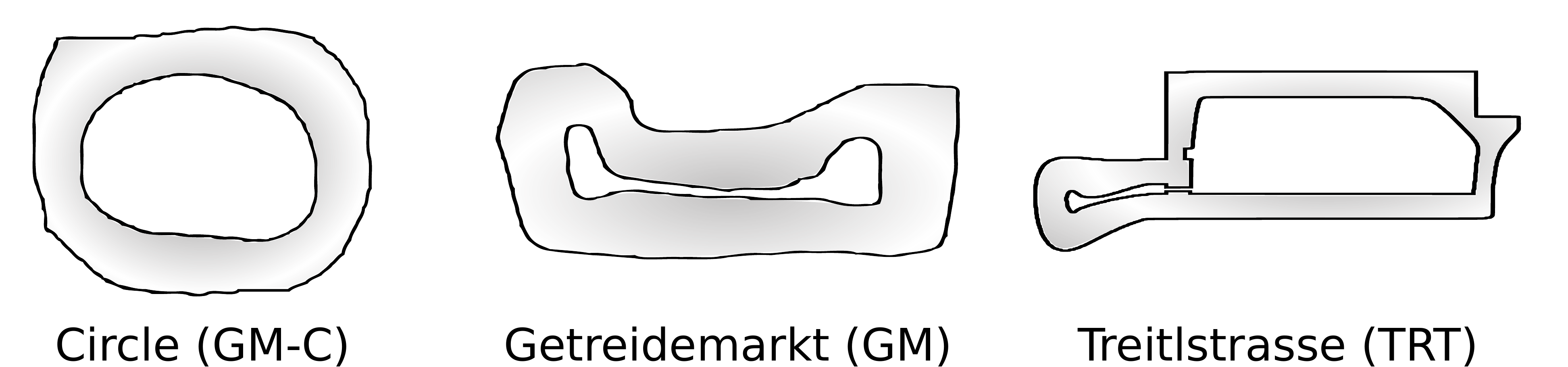}
    \caption{
        Different racing tracks used for the Safe Driving and Follow Lead Vehicle tasks. 
        The tracks were first physically created at our lab and the map imported into the simulator. 
        The Safe Driving task was trained on both GM ($21.25\,m$) and TRT ($51.65\,m$), and physically deployed in GM. 
        The Follow Lead Vehicle task was trained and deployed on GM-C ($13.50\,m$). 
    }
    \label{fig:racing_tracks}
\end{figure*}

As mentioned in the main text, the objective of this task is to complete 1 lap around the track in a safe manner. The requirements are in  found in Table \ref{tab:requirements} of the main text, and the parameters used can be found in Table \ref{tab:safedriving_parameters}. 
The training has been carried on in simulation using three different tracks. The cars' starting positions and the simulation parameters has been randomly sampled at the beginning of each episode. 
The tracks reported in Figure \ref{fig:racing_tracks} have been physically created in our facilities at the Technical University of Vienna:  Getreidemarkt Circle (GM-C), Getreidemarkt (GM), and Treitlstrasse (TRT). 
For the accompanying video, we show the single-agent behavior in the GM circuit, and the follow-leading vehicle in GM-C.

\subsection{Follow Lead Vehicle}
\begin{figure*}[ht]
    \centering
    \includegraphics[width=0.8\linewidth]{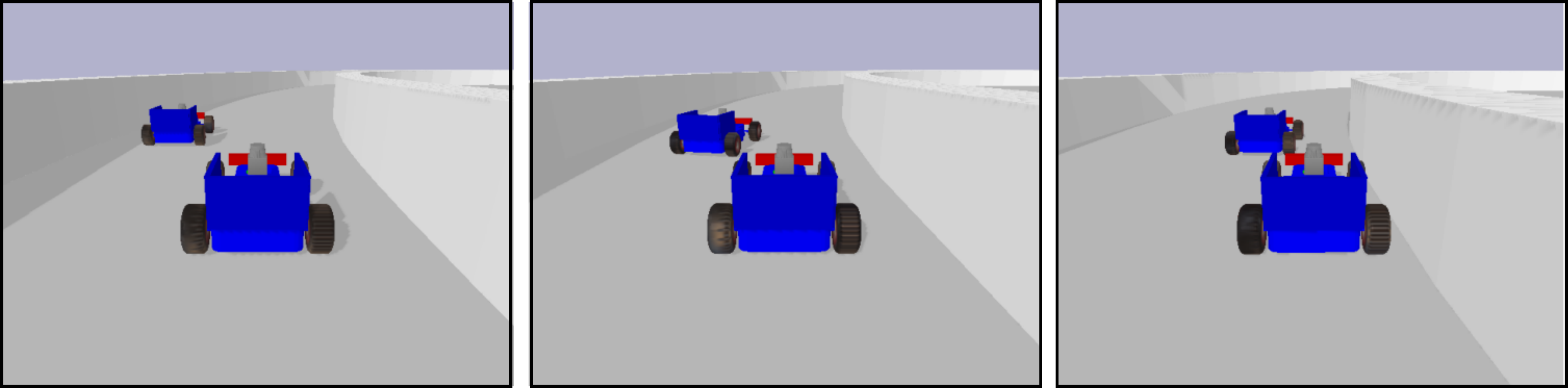}
    \caption{Racecar environment used in Safe Driving and Follow Lead Vehicle.}
    \label{fig:racecar_storyboard}
\end{figure*}

The second driving task uses the same environment, shown in Figure \ref{fig:racecar_storyboard}. 
The agent's objective is also to complete a single lap around a track, but this time it has to do so while following a leading vehicle. In this case, the comfort requirements are: encouraging a small steering angle ($\alpha$), encouraging smooth controls ($|a|$), and encouraging the agent to keep a distance between $[d^{\emph{lead}}_{\text{min}, \emph{comf}}, d^{\emph{lead}}_{\text{max}, \emph{comf}}]$. In contrast to the safe driving example, note that here there are no comfort requirements to keep a specific speed, nor are there comfort requirements for driving in the middle of the track.  The requirements are found in Table \ref{tab:followlead_requirements}, and the parameters used can be found in Table \ref{tab:followlead_parameters}. The map used for this task is the Getreidemarkt circle (GMC), observed in Figure \ref{fig:racing_tracks}.

\begin{table}[t]
    \label{tab:followlead_requirements}
    \centering
    \begin{tabular}{|l|l|l|l|}
    \hline
    Req Id & Formula Id & Formula \\
    \hline \hline
    Req1 & $\varphi_1$ & $\texttt{achieve } L(s) = 1.0$ \\
    Req2 & $\varphi_2$ & $\texttt{ensure } d_{\emph{walls}}(s) > 0$ \\
    Req4 & $\varphi_3$ & $\texttt{encourage } d_{\emph{lead}}(s) \geq d^{\emph{lead}}_{\text{min}, \emph{comf}}$ \\
    Req5 & $\varphi_4$ & $\texttt{encourage } d_{\emph{lead}}(s) \leq d^{\emph{lead}}_{\text{max}, \emph{comf}}$ \\
    Req6 & $\varphi_5$ & $\texttt{encourage } |\alpha| \leq \alpha_{\emph{comf}}$ \\
    Req7 & $\varphi_6$ & $\texttt{encourage } |a| \leq \Delta a$ \\
    \hline
    \end{tabular}
\caption{Follow Lead Vehicle -- formalized requirements}
\end{table}

\subsection{Lunar-Lander with Obstacle:}
The lunar lander's objective 
is to land at the pad with coordinates 
$(0,0)$. In this example, we assume infinite 
fuel. Landing outside of the pad is also 
possible (as long as the impact velocity does not exceed a threshold).
We allow continuous actions that 
allow firing the engine to the left or right, 
firing the main engine, and doing nothing. 
We add an obstacle to the 
environment, that makes the landing harder.
The original state space of the lunar lander
is the tuple $(x, y, \dot{x}, \dot{y}, \theta, \dot{\theta})$, where $(x,y)$ is the position of the 
lander, $(\dot{x},\dot{y})$ is its velocity, $\theta$ 
is the angle of the lander with respect to 
the direction of gravity ($y$-axis), and $\dot{\theta}$ is 
its angular velocity. The lander also has contact sensors for both of its legs, and we further enhance the state space with the obstacle coordinates (left, right, top, bottom) and a flag variable on whether the pole has collided or not.

\begin{figure*}[ht]
    \centering
    \includegraphics[width=\linewidth]{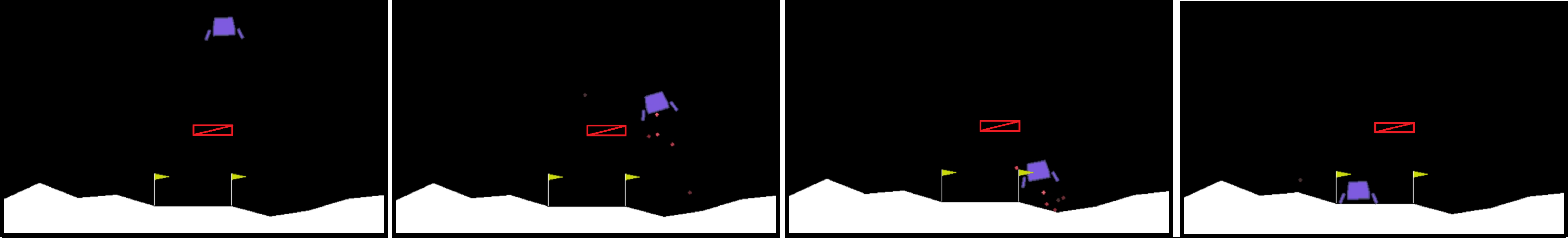}
    \caption{Lunar lander overcomes the obstacle.}
    \label{fig:lunarlander_storyboard}
\end{figure*}

We introduce a maximum number of steps to explicitly define an episode.
The obstacle size and position are fixed to be above the target area and force the lander to avoid the obstacle to reach the goal. 
The used parameters are found in Table \ref{tab:lunarlander_parameters}. Figure \ref{fig:lunarlander_storyboard} shows some frames of the lander surrounding the obstacle (red) to safely reach its target.

To define the requirements, we use the following constants: 
(1) $G$ -- the coordinates of the landing area, 
(2) $O$ -- the area that is occupied 
by the static obstacle, 
(3) $x_{\emph{lim}}$ -- the limit of the world, 
(4) $\theta_{\emph{comf}}$ -- the maximum 
comfortable angle and (5) $\dot{\theta}_{\emph{comf}}$ -- the maximum
comfortable angular velocity.
Table~\ref{tab:ll} lists 
the requirements formalized in the proposed specification language.

\begin{table}[ht]
\centering
    \begin{tabular}{|l|l|l|l|}
    \hline
    Req Id & Formula Id & Formula\\
    \hline \hline
    $\text{Req}_1$  & $\varphi_3$ & $\texttt{conquer } d(\mathbf{r},G) = 0$ \\
    $\text{Req}_2$  & $\varphi_1$ & $\texttt{ensure } d(\mathbf{r},O) \geq 0$  \\
    $\text{Req}_3$  & $\varphi_2$ & $\texttt{ensure } |x|\leq x_{\emph{lim}}$  \\
    $\text{Req}_4$  & $\varphi_4$ & $\texttt{encourage } |\theta| \leq \theta_{\emph{comf}}$ \\
    $\text{Req}_5$  & $\varphi_5$ & $\texttt{encourage } |\dot{\theta}| \leq \dot{\theta}_{\emph{comf}}$  \\
    \hline
    \end{tabular}
    \caption{Lunar lander with obstacle example --  formalized requirements.}
    \label{tab:ll}
\end{table}

\subsection{Bipedal-Walker:}
\begin{figure*}[ht]
    \centering
    \includegraphics[width=\linewidth]{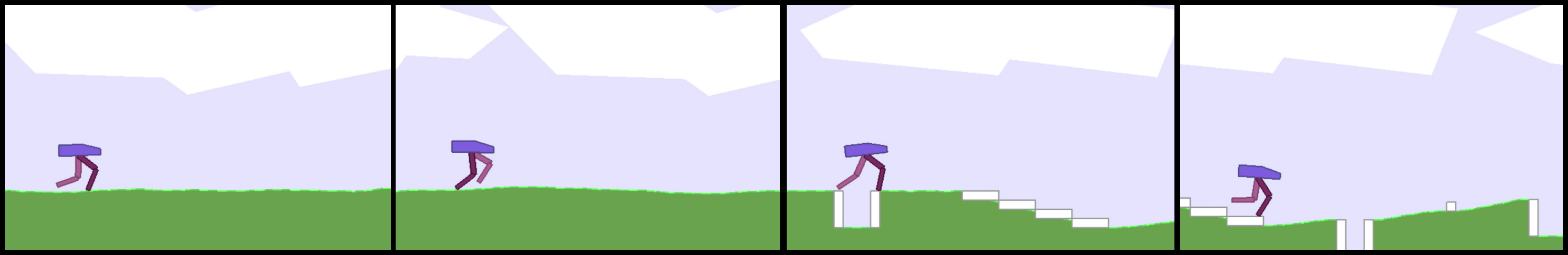}
    \caption{Left two frames: The normal bipedal walker. Right two: The hardcore version with obstacles and holes.}
    \label{fig:bipedalwalker_storyboard}
\end{figure*}

In this case study, the main objective for 
the robot is to move forward without falling.
We consider two variants of this case study -- 
the {\em classical} one with the flat
terrain and 
the {\em hardcore} one with holes and obstacles.
A state in this case study is the tuple 
$(\dot{x}, \dot{y}, \theta, \dot{\theta}, \textbf{l})$, where $\dot{x}$ is the horizontal velocity, 
$\dot{y}$ is the vertical velocity, 
$\theta$ is the hull angle, 
$\dot{\theta}$ is the angular velocity 
and $\textbf{l}$ is a vector of $10$ 
LiDAR range-finder measurements. 
In the original environment, there are additional variables in the agent's observation (e.g., joints position, joints angular speed), and we omit their definition because not used in the formalization.
We did not alter the agent's observation space for this environment. 
Since we wanted to showcase the \texttt{achieve} specification, the environment terminates upon reaching the goal.
Table \ref{tab:bipedalwalker_parameters} (end of document) lists the used parameters. Figure \ref{fig:bipedalwalker_storyboard} shows two frames of the normal environment, and two frames of the hardcore environment.

We define the following constants: 
(1) $O$ -- the set of coordinates occupied 
by the static obstacle, 
(2) $\theta_{\emph{comf}}$ -- the maximum 
comfortable angle, (3) $\dot{\theta}_{\emph{comf}}$ -- the maximum 
comfortable angular velocity, and
(4) $\dot{y}_{\emph{comf}}$ -- the maximum 
comfortable vertical velocity.
Table~\ref{tab:bw} lists 
the informal requirements collected for the 
bipedal walker example, and their formalization 
in STL.

\begin{table}[ht]
\centering
    \begin{tabular}{|l|l|l|l|}
    \hline
    Req Id & STL Id &  Formula \\
    \hline \hline
    $\text{Req}_2$  & $\varphi_2$ & $\texttt{achieve } d(x,G) = 0 $  \\
    $\text{Req}_2$  & $\varphi_1$ & $\texttt{ensure } \min_{l \in \textbf{l}} d(l, O) > 0$  \\
    $\text{Req}_3$  & $\varphi_3$ & $\texttt{encourage } |\theta| \leq \theta_{\emph{comf}}$  \\
    $\text{Req}_4$  & $\varphi_4$ & $\texttt{encourage } |\dot{\theta}| \leq \dot{\theta}_{\emph{comf}}$  \\
    $\text{Req}_5$  & $\varphi_5$ & $\texttt{encourage } |\dot{y}| \leq \dot{y}_{\emph{comf}}$  \\
    \hline
    \end{tabular}
    \caption{Bipedal-walker example -- formalized requirements.}
    \label{tab:bw}
\end{table}

\newpage
\twocolumn

\subsection{Tables of Environment Parameters}

\begin{table}[ht]
\begin{center}
\begin{tabular}{ |m{16.5em} m{4em}| }
\hline
\multicolumn{2}{|l|}{\textbf{Safe Driving}} \\
\hline
\multicolumn{2}{|c|}{\textbf{Episode Conditions}} \\
\hline
max\_steps & $2500$   \\ 
\multicolumn{2}{|m{20em}|}{Steps before episode termination.} \\ \hline
track  & GM  \\ 
  & TRT  \\ 
\multicolumn{2}{|m{20em}|}{ Track maps used for training, randomized.} \\ \hline
frame\_skip & $10$   \\ 
\multicolumn{2}{|m{20em}|}{Number of frames to skip after each action.} \\ \hline
terminate\_on\_collision  & True  \\ 
\multicolumn{2}{|m{20em}|}{ Episode ends if vehicle collides with walls.} \\ \hline
dt  & $0.01$  \\ 
\multicolumn{2}{|m{20em}|}{ Simulation integration time (s).} \\ \hline

\multicolumn{2}{|c|}{\textbf{Reward Parameters}} \\
\hline
comf\_wall\_dist & $0.5$ \\
\multicolumn{2}{|m{20em}|}{ Normalized desired distance to walls.} \\ \hline
comf\_min\_vel & $2.0$ \\
\multicolumn{2}{|m{20em}|}{ Desired minimum speed (m/s).} \\ \hline
comf\_max\_vel & $3.5$ \\
\multicolumn{2}{|m{20em}|}{ Desired maximum speed (m/s).} \\ \hline
comf\_max\_steering & $0.1$ \\
\multicolumn{2}{|m{20em}|}{ Target normalized max steering absolute value.} \\ \hline
comf\_max\_action\_norm & $0.25$ \\
\multicolumn{2}{|m{20em}|}{ $l_2$-norm deviation on consecutive actions.} \\ \hline

\multicolumn{2}{|c|}{\textbf{Observation Parameters}} \\
\hline
use\_history\_wrapper & True \\
\multicolumn{2}{|m{20em}|}{ Stack previous actions and observations.} \\ \hline
n\_last\_actions & $3$ \\
\multicolumn{2}{|m{20em}|}{ $N$ actions to stack with action-history wrapper.} \\ \hline
n\_last\_obersvations & $1$ \\
\multicolumn{2}{|m{20em}|}{ $N$ observations to stack with obs-history wrapper.} \\ \hline
obs\_names & lidar\_64  \\
 &  velocity\_x \\
\multicolumn{2}{|m{20em}|}{ Observation data for agent.} \\ \hline
\end{tabular}
\end{center}
\caption{Parameters used for the safe driving environment.}
\label{tab:safedriving_parameters}
\end{table}

\begin{table}[ht]
\begin{center}
\begin{tabular}{ |m{17em} m{3.5em}| }
\hline
\multicolumn{2}{|l|}{\textbf{Follow Lead Vehicle}} \\
\hline
\multicolumn{2}{|c|}{\textbf{Episode Conditions}} \\
\hline
max\_steps & $3000$   \\ 
\multicolumn{2}{|m{20em}|}{Steps before episode termination.} \\ \hline
track  & GMC  \\ 
\multicolumn{2}{|m{20em}|}{ Track maps used for training.} \\ \hline

\multicolumn{2}{|c|}{\textbf{Reward Parameters}} \\
\hline
comf\_wall\_dist & $0.5$ \\
\multicolumn{2}{|m{20em}|}{ Normalized desired distance to walls.} \\ \hline
comf\_min\_dist\_lead & $-2.0$ \\
\multicolumn{2}{|m{20em}|}{ Desired minimum distance to leader (m).} \\ \hline
comf\_max\_dist\_lead & $-1.5$ \\
\multicolumn{2}{|m{20em}|}{ Desired maximum distance to leader (m).} \\ \hline
comf\_max\_steering & $0.1$ \\
\multicolumn{2}{|m{20em}|}{ Target normalized max steering absolute value.} \\ \hline
comf\_max\_action\_norm & $0.25$ \\
\multicolumn{2}{|m{20em}|}{ $l_2$-norm deviation on consecutive actions.} \\ \hline

\multicolumn{2}{|c|}{\textbf{Observation Parameters}} \\
\hline
n\_last\_obersvations & $3$ \\
\multicolumn{2}{|m{20em}|}{ $N$ observations to stack with obs-history wrapper.} \\ \hline

\multicolumn{2}{|c|}{\textbf{Leading Vehicle Parameters}} \\
\hline
algorithm & ftg \\
\multicolumn{2}{|m{20em}|}{ ``Follow the Gap" driving controller.} \\ \hline
gap\_threshold & $2.0$ \\
\multicolumn{2}{|m{20em}|}{ Parameter for ftg algorithm, gap size (m).} \\ \hline
lead\_min\_speed & $1.25$ \\
\multicolumn{2}{|m{20em}|}{ Random leader speed (m/s) min value.} \\ \hline
lead\_max\_speed & $1.75$ \\
\multicolumn{2}{|m{20em}|}{ Random leader speed (m/s) max value.} \\ \hline
lead\_observation & lidar\_64 \\
\multicolumn{2}{|m{20em}|}{Observation data for leader. } \\ \hline

\end{tabular}
\end{center}
\caption{Parameters used for the Follow Lead Vehicle environment. Non stated episode conditions and observation parameters are the same as the Safe driving example. }
\label{tab:followlead_parameters}
\end{table}

\begin{table}[ht]
\begin{center}
\begin{tabular}{ |m{18em} m{2.5em}| }
\hline
\multicolumn{2}{|l|}{\textbf{Lunar lander with obstacle}} \\
\hline
\multicolumn{2}{|c|}{\textbf{Episode Conditions}} \\
\hline
max\_steps & $600$   \\ 
\multicolumn{2}{|m{20em}|}{Steps before episode termination.} \\ \hline
terminate\_on\_collision  & True  \\ 
\multicolumn{2}{|m{20em}|}{ Episode ends if lander collides with ground or with obstacle.} \\ \hline
x\_limit  & $1.0$  \\ 
\multicolumn{2}{|m{20em}|}{ Episode ends if $|x| \geq x_{\text{limit}}$. \newline
Note: there is no $y_\text{limit}$.}\\ \hline
\multicolumn{2}{|c|}{\textbf{Reward Parameters}} \\
\hline
x\_target & $0.0$   \\ 
y\_target & $0.0$   \\
\multicolumn{2}{|m{20em}|}{The lander's ideal goal.} \\ \hline
x\_target\_tol & $0.15$   \\ 
\multicolumn{2}{|m{20em}|}{Goal area.\newline
$G=\{(x,0): |x - x_{\text{target}}| \leq x_{\text{target\_tol}}\}$} \\ \hline
theta\_target & $0.0$   \\ 
\multicolumn{2}{|m{20em}|}{The lander's most comfortable angle.} \\ \hline
theta\_comf & $\pi/3$   \\ 
\multicolumn{2}{|m{20em}|}{Comfort limit. $|\theta - \theta_{\text{target}}| \leq \theta_{\text{comf}} $} \\ \hline
theta\_dot\_target & $0.0$   \\ 
\multicolumn{2}{|m{20em}|}{The lander's most comfortable angular velocity.} \\ \hline
theta\_dot\_comf & $0.5$   \\ 
\multicolumn{2}{|m{20em}|}{Comfort limit. $|\dot{\theta} - \dot{\theta}_{\text{target}}| \leq \dot{\theta}_{\text{comf}} $} \\ \hline
\multicolumn{2}{|c|}{\textbf{Initial Conditions}} \\
\hline
x\_offset & $0.1$\\
\multicolumn{2}{|m{20em}|}{Lander's starting $x$-position,   $x_{0}\in \pm x_\text{offset}$ } \\ \hline
obstacle\_left & $0.0$ \\
obstacle\_right & $0.2$ \\
obstacle\_bottom & $0.53$ \\
obstacle\_top & $0.46$ \\
\multicolumn{2}{|m{20em}|}{Obstacle size and position is fixed.} \\ \hline
\end{tabular}
\end{center}
\caption{Parameters used for the lunar lander with obstacle environment. Non-stated parameters were not altered. Angles given in radians. Spatial coordinates normalized with default environment resolution.}
\label{tab:lunarlander_parameters}
\end{table}

\begin{table}[ht]
\begin{center}
\begin{tabular}{ |m{18em} m{2.5em}| }
\hline
\multicolumn{2}{|l|}{\textbf{Bipedal Walker}} \\
\hline
\multicolumn{2}{|c|}{\textbf{Episode Conditions}} \\
\hline
max\_steps & $500$   \\ 
\multicolumn{2}{|m{20em}|}{Steps before episode termination.} \\ \hline
terminate\_on\_collision  & True  \\ 
\multicolumn{2}{|m{20em}|}{ Episode ends if walker's hull makes contact with the ground or an obstacle.} \\ \hline
\multicolumn{2}{|c|}{\textbf{Reward Parameters}} \\
\hline
lidar\_offset & $0.225$ \\
\multicolumn{2}{|m{20em}|}{Since the LiDAR origin is inside the hull, a constant offset is considered. $\textbf{l} \leftarrow \textbf{l} - l_{\text{offset}}$ } \\ \hline
theta\_target & $0.0$   \\ 
\multicolumn{2}{|m{20em}|}{The hull's most comfortable angle.} \\ \hline
theta\_comf & $0.0873$   \\ 
\multicolumn{2}{|m{20em}|}{Comfort limit. $|\theta - \theta_{\text{target}}| \leq \theta_{\text{comf}} $} \\ \hline
theta\_dot\_target & $0.0$   \\ 
\multicolumn{2}{|m{20em}|}{The hull's most comfortable angular velocity.} \\ \hline
theta\_dot\_comf & $0.25$   \\ 
\multicolumn{2}{|m{20em}|}{Comfort limit. $|\dot{\theta} - \dot{\theta}_{\text{target}}| \leq \dot{\theta}_{\text{comf}} $} \\ \hline
y\_dot\_target & $0.0$   \\ 
\multicolumn{2}{|m{20em}|}{The hull's most comfortable $y$-velocity.} \\ \hline
y\_dot\_comf & $0.1$   \\ 
\multicolumn{2}{|m{20em}|}{Comfort limit. $|\dot{y} - \dot{y}_{\text{target}}| \leq \dot{y}_{\text{comf}} $} \\ \hline
\end{tabular}
\end{center}
\caption{Parameters used for the bipedal walker environments Non-stated parameters were not altered. Angles given in radians. Spatial coordinates normalized with default environment resolution.}
\label{tab:bipedalwalker_parameters}
\end{table}

\end{document}